\documentclass[10pt,twocolumn,letterpaper]{article}
\usepackage{cvpr}      
\usepackage{times}  
\usepackage{helvet}  
\usepackage{courier}  
\usepackage[utf8]{inputenc}
\usepackage[T1]{fontenc}    
\usepackage{inconsolata}

\usepackage[hyphens]{url}  
\usepackage{booktabs}
\usepackage{multirow}
\usepackage{adjustbox}
\usepackage{wrapfig}
\usepackage{colortbl}      
\usepackage{caption} 

\usepackage{natbib}  
\usepackage{latexsym}

\usepackage{algorithm}
\usepackage{algpseudocode}
\algrenewcommand\algorithmicrequire{\textbf{Input:}}
\algrenewcommand\algorithmicensure{\textbf{Output:}}
\algnewcommand{\LineComment}[1]{\State $\triangleright$ #1}
\algrenewcommand\algorithmiccomment[1]{\hfill$\triangleright$ #1}

\usepackage{amsfonts}       
\usepackage{nicefrac}       
\usepackage{microtype}      
\usepackage{empheq}

\usepackage{amsthm}

\usepackage{enumerate}
\usepackage{mathrsfs }
\usepackage{makecell}

\usepackage{mathtools}

\usepackage{listings}
\usepackage{xspace}
\newcommand{\method}{{\textup{\textsc{Visual-Subtitle Integration}}}\xspace}
\newcommand{\methods}{{\textup{\textsc{VSI}}}\xspace}
\newcommand{\lvb}{{\textup{\textsc{LongVideoBench}}}\xspace}
\newcommand{\videomme}{{\textup{\textsc{Video-MME}}}\xspace}

\definecolor{cvprblue}{rgb}{0.21,0.49,0.74}
\usepackage[pagebackref,breaklinks,colorlinks,allcolors=cvprblue]{hyperref}

\def\paperID{9004} 
\def\confName{CVPR}
\def\confYear{2026}
\title{VSI: Visual–Subtitle Integration for Keyframe Selection to Enhance \\Long Video Understanding}
\author{
   Jianxiang He \textsuperscript{1,*},
   Meisheng Hong \textsuperscript{2,*},
   Jungang Li \textsuperscript{1},
   Weiyu Guo \textsuperscript{1},
   Xuming Hu \textsuperscript{1,\textdagger},
   Hui Xiong \textsuperscript{1,\textdagger}\\
   \textsuperscript{1} AI Thrust, HKUST(GZ), \textsuperscript{2} Shandong University \\
   \textsuperscript{*} These authors contributed equally to this work. \\
   \textsuperscript{\textdagger} Corresponding authors.
}

\begin{document}
\maketitle

\begin{abstract}
Multimodal large language models (MLLMs) demonstrate exceptional performance in vision-language tasks, yet their processing of long videos is constrained by input context length and high computational costs. Sparse frame sampling thus becomes a necessary preprocessing step, with sampled frame quality directly impacting downstream performance. Existing keyframe search algorithms achieve a balance between efficiency and sampled frame quality but heavily rely on the visual modality alone. This makes them difficult to adapt to text-related tasks and often leads to retrieval results deviating from core semantic content. To address this, we propose the \method(\methods), a multimodal keyframe retrieval framework. It employs a dual-branch collaborative retrieval approach combining Video Search and Subtitle Match to fuse complementary visual and textual information for precise localization. Experiments on LongVideoBench and VideoMME demonstrate that \methods achieves state-of-the-art accuracy in keyframe retrieval while delivering breakthrough performance in text-related tasks and exhibiting strong generalization across other tasks. The source code is available at: \url{https://github.com/Jacksonha7/Visual-Subtitle-Integration.git}
\end{abstract}

\vspace{-1.75em}
\section{Introduction}
\begin{figure}[t]
\centering
\includegraphics[width=0.95\linewidth]{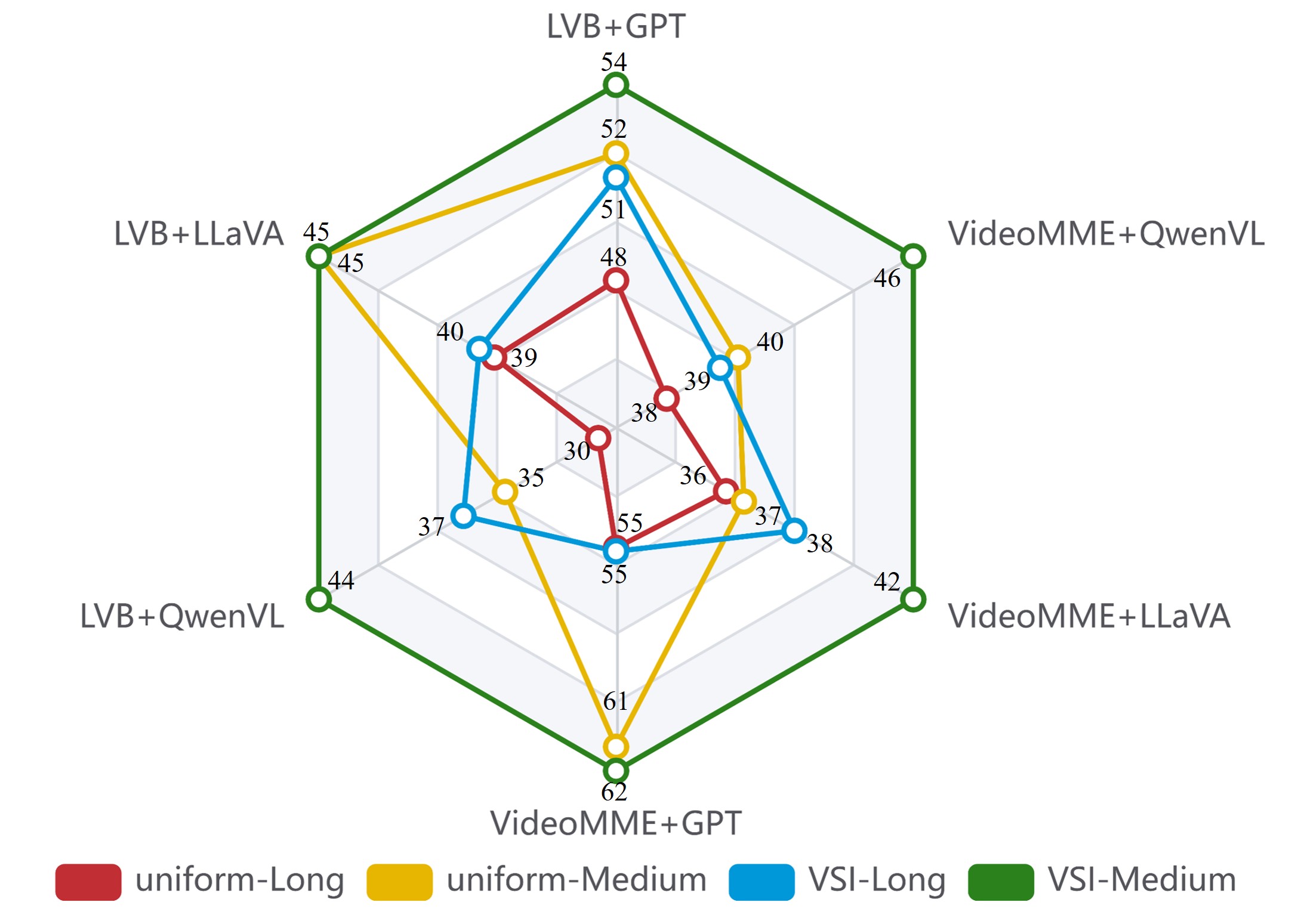}
\vspace{-0.7em}
\caption{\textbf{Comparison with Baseline}. In medium and long video settings of \lvb and \videomme, \methods generally outperforms uniform-sampling baselines and shows clear gains over GPT-4o, LLaVA-Video-7B-Qwen2, and Qwen2.5-VL-7B-Instruct in most settings.}
\label{fig1}
\vspace{-1.5em}
\end{figure}
Multimodal Large Language Models (MLLMs) have demonstrated exceptional capabilities across a diverse range of vision-language tasks, including image captioning, Video Question Answering (VideoQA), and Optical Character Recognition (OCR). These models achieve such strong performance by encoding visual information into model-compatible input representations. Upon extension to the video domain, video-based MLLMs have attracted considerable research interest in recent years. However, due to the inherent complex spatio-temporal dynamics of video data, which is different from that of static images, the processing of video data faces great challenges.

Constrained by computational resources and inherent model capabilities, existing MLLMs typically employ sparse frame sampling when processing video. However, different sampling strategies can lead to significant variations in video content analysis results, ultimately impacting MLLM performance on downstream tasks. Therefore, the core challenge lies in determining the optimal frame sampling strategy to maximize MLLM video understanding accuracy while minimizing computational overhead. Most existing MLLMs (e.g., LLaVa-Video~\citep{36}) employ uniform frame sampling for video processing, which frequently omits query-relevant critical information. Current research focuses on exploring keyframe extraction methods that require no training. For instance, LongVU~\citep{longvu} utilizes pre-trained feature extractors (like DINOv2-1B) to identify frames exhibiting significant cross-frame differences; CoS~\citep{cos} employs LLaVA-1.5-13B to filter query-relevant frames as input for Video-MLLM, though this approach introduces substantial computational overhead. AKS~\citep{aks} performs adaptive keyframe selection by comprehensively evaluating the relevance of keyframes to prompts and their coverage within the video. Another class of keyframe search algorithms attempts to replace MLLM processing of initial video frames with lightweight object detection models to screen keyframes for downstream video question-answering tasks. \textsc{Tstar}~\citep{ye2025rethinking} represents this approach by identifying key objects and boosting the sampling weights of frames containing them. Building upon this foundation, Visual Semantic Logic Search(\textsc{VSLS})~\citep{vsls} explicitly models relationships between video frames to mitigate the issue of missing temporal information. The aforementioned frameworks provide effective solutions for minimizing computational overhead while maintaining the accuracy of MLLM-based video understanding.

However, current keyframe search algorithms still face two fundamental challenges: (1) Limited task types can be handled. In VideoQA tasks, existing methods exhibit significant limitations: they can only effectively improve performance on visually-strong sub-tasks, while showing minimal improvement for text-strong sub-tasks, making it difficult to meet performance demands in multimodal scenarios; (2) Failure to make full use of multimodal information. Existing approaches rely solely on visual monomodal retrieval, lacking targeted guidance from modalities like text. This leads to keyframes being overly focused on visually dense regions, shifting the emphasis away from true semantic content and consequently resulting in low search accuracy.

To address the aforementioned challenges, we propose \method(\methods). This framework employs a dual-branch collaborative retrieval mechanism comprising Video Search and Subtitle Match: the Video Search branch performs initial keyframe sampling based on visual feature extraction and object detection, while the Subtitle Match branch obtains complementary textual information through semantic similarity calculations. Outputs from both branches undergo dynamic fusion strategies to achieve deep cross-modal information interaction. The core advantages of this dual-branch collaborative design manifest in two aspects: First, through the targeted guidance of textual modality information, \methods effectively handles the VideoQA subtask dominated by textual relevance, overcoming the scenario limitations of existing unimodal approaches and significantly expanding the applicability of keyframe retrieval algorithms. Second, the deep fusion of cross-modal information enables the framework to precisely locate keyframes containing core semantics, avoiding retrieval bias caused by excessive focus on visual information.

We comprehensively evaluated \methods on several public benchmarks, including \lvb~\citep{lvb}, a large-scale benchmark for long video understanding, and \videomme~\citep{videomme}, a widely used dataset for multimodal video question answering. 
Experimental results show that VSI consistently outperforms uniform sampling on \lvb and improves performance in most settings on \videomme, while also surpassing unimodal keyframe-search baselines on the full, image-only, and text-only subsets of \lvb. Notably, on text-related tasks in \lvb (Table~\ref{ablation_text_ptonly}), VSI improves keyframe search accuracy (Acc) from 29.48 to \textbf{45.00} under the 8-frame setting. Furthermore, when keyframes selected by \methods are fed into GPT-4o~\citep{hurst2024gpt} for Long-VideoQA, the Long split score improves from 53.76 to \textbf{69.57}.

Our principal contributions are threefold:
\begin{itemize}
    \item We extend keyframe retrieval from \textbf{single-modality search} to \textbf{dual-branch multimodal search}, enabling robust handling of text-related queries while maintaining strong performance on other task types.
    \item We propose a \textbf{plug-and-play} method that requires no additional training and is both lightweight and flexible, with search models and encoding models that can be flexibly replaced according to different needs.
    \item We conduct extensive empirical evaluations across multiple datasets. Results validate both effectiveness and efficiency, with VSI achieving \textbf{73.89\%} keyframe-search accuracy on \lvb while maintaining competitive computational cost.
\end{itemize}
\section{Method}
\begin{figure*}[ht]
\centering
\includegraphics[width=\textwidth]{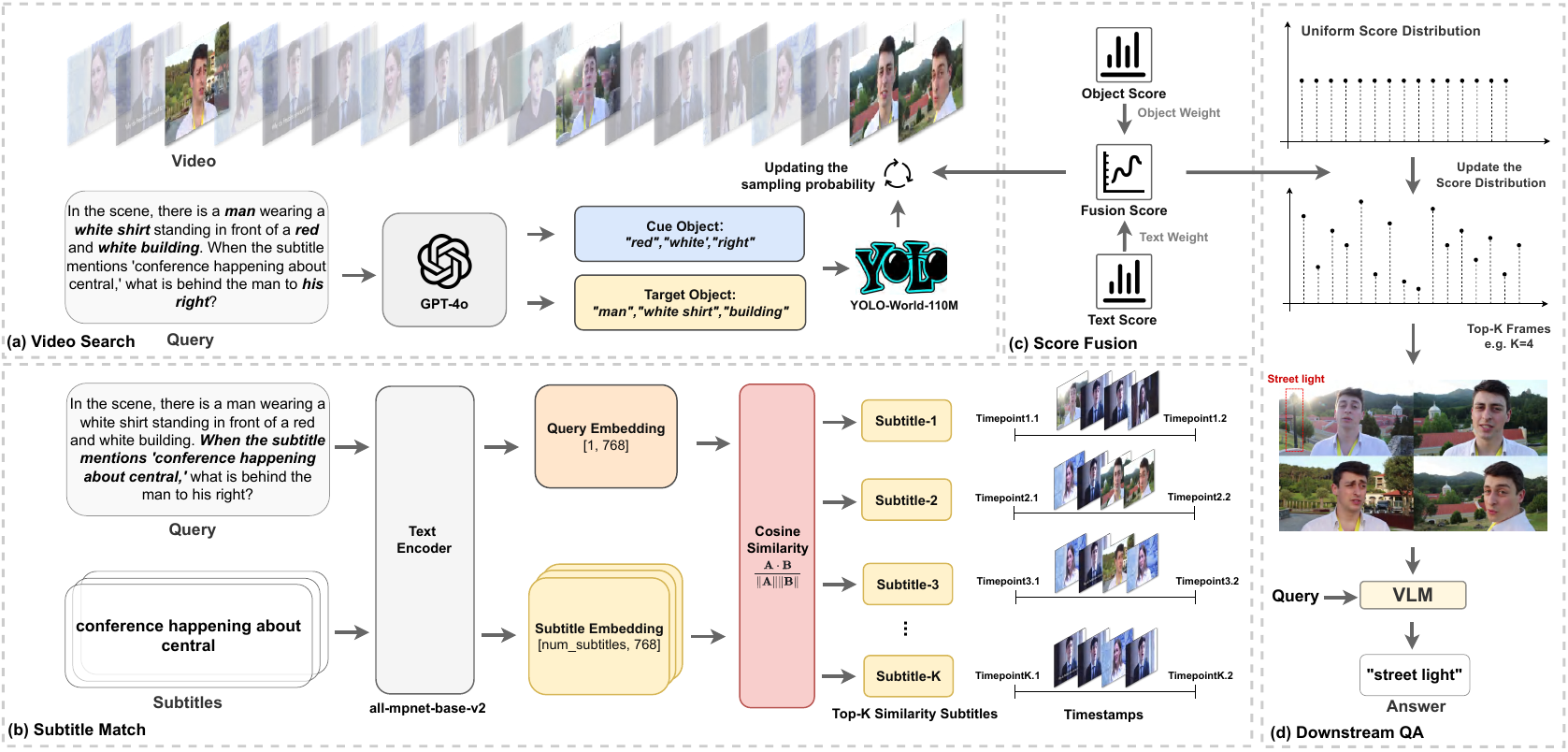}
\caption{\textbf{Framework for \method}. The dual-branch architecture comprises: (a) \textbf{Video Search branch} leveraging YOLO-World to identify query-relevant objects; (b) \textbf{Subtitle Match branch} employing contrastive embeddings to retrieve subtitle-matching segments; (c) Confidence scores from both modalities are fused to update frame-wise relevance probabilities through spline interpolation. After the iteration, high-confidence frames were subsequently propagated to downstream QA tasks. The figure shows a complete real example from keyframe search to the completion of a VideoQA task.}
\label{fig:framework}
\end{figure*}
This paper proposes a \method (\methods) mechanism that integrates visual-temporal and textual semantic retrieval to overcome unimodal limitations. It employs a dual-branch collaborative mechanism: (1) The \textbf{video search branch} samples keyframes based on query-relevant objects via object detection and assigns them high weights; (2) The \textbf{subtitle matching branch} enhances corresponding frame weights through query-caption semantic similarity. After initializing frame weights, the results fuse scores from both branches to update sampling probabilities, guiding sampling toward semantically dense regions. The top-$k$ keyframes are selected based on the score distribution.
\subsection{Task Formulation}
Given a video sequence $V = \{v_i\}_{i=1}^{T}$ consisting of $T$ frames and a textual query $Q$, the keyframe retrieval task aims to identify the most concise keyframe subset $\mathcal{K} = \{v_{k_j}\}_{j=1}^{K} \subset V$ that satisfies two criteria: 
1)~\textit{Answer Preservation}: The extracted subset $\mathcal{K} \subset V$ must guarantee answer equivalence, i.e., $F(\mathcal{K}, Q) \sim F(V, Q)$, where $F(\cdot)$ denotes the video question-answering model;  
2)~\textit{Minimality}: $\mathcal{K}$ should form an irreducible subset that preserves the correctness of the answer, with any frame removal altering the model's response, while maintaining efficiency.

\subsection{Visual-Subtitle Integration}
\subsubsection{Frame Sampling.}
To expedite the search procedure, we employ a selective sampling approach rather than processing every video frame. Let $N_{\text{video}}$ represent the total frame count, and $D$ be an initial uniform sampling distribution across all frames. The frame selection is performed as:
\begin{equation}
    I_{\text{sel}} = \mathcal{R}(D \circ N_{\text{video}}, n_{\text{sample}}),
\end{equation}
where $\mathcal{R}(\cdot, n_{\text{sample}})$ randomly picks $n_{\text{sample}}$ frames following the weighted distribution $D \circ N_{\text{video}}$. To further leverage the detector's detection capabilities, we arrange the selected frames into a $m \times m$ matrix format, which necessitates:
\begin{equation}
n_{\text{sample}} = m^2, \quad m \in \mathbb{N}^+, \ \text{and} \ n_{\text{sample}} \ll N_{\text{video}}.
\end{equation}
During the initial iteration, uniformly spaced sampling is employed to ensure comprehensive coverage of the video, establishing a preliminary understanding of its content while avoiding excessive early focus on any single region. Subsequently, $D$ adjusts through multiple sampling rounds to concentrate on frames of greater interest within the video.

\subsubsection{Object Detection and Scoring.}

To provide initial visual grounding, \methods prompts the VLM to scan a set of evenly spaced frames and identify relevant objects from two semantic categories:  
1)~\textit{Target Objects} $T_{\text{obj}}$: key visual entities directly related to answering the query;  
2)~\textit{Cue Objects} $C_{\text{obj}}$: contextual entities that offer indirect clues, such as common spatial configurations or environmental hints. Cue objects provide contextual clues to identify target object-related potential scenarios and can pinpoint relevant regions even in the absence of detected target objects, thereby boosting search efficiency. The specific prompt is provided in the appendix. We formalize this as:
\begin{equation}
T_{\text{obj}},\, C_{\text{obj}} := \text{VLMs}(V, Q, K), \quad 
\mathcal{T} := T_{\text{obj}} \cup C_{\text{obj}}.
\end{equation}

During each sampling iteration, we apply an efficient detector (e.g., \textsc{YOLO-World}~\citep{yolo}) to the selected $n_{\text{sample}}$ frames. Each frame is evaluated by checking the overlap between detected objects and the predefined target set $\mathcal{T}$. Let $\mathcal{O}_t$ denote the set of objects detected in frame $t$, $s_o$ the detection confidence of object $o \in \mathcal{O}_t$, and $w_o$ the task-specific importance weight for object $o$. The object-based frame score is defined as: 
\begin{equation}
S_{\text{obj}}(t) = \max_{o \in \mathcal{O}_t \cap \mathcal{T}} \left( s_o \cdot w_o \right).
\end{equation}

If $S_{\text{obj}}(t)$ exceeds a threshold $\tau_{\text{obj}}$, the corresponding frame is added to a priority queue for downstream processing. This mechanism ensures that visually salient and semantically aligned frames are prioritized for further analysis. 

\subsubsection{Subtitle Matching}
\begin{figure*}[hbtp]
\centering
\includegraphics[width=\textwidth]{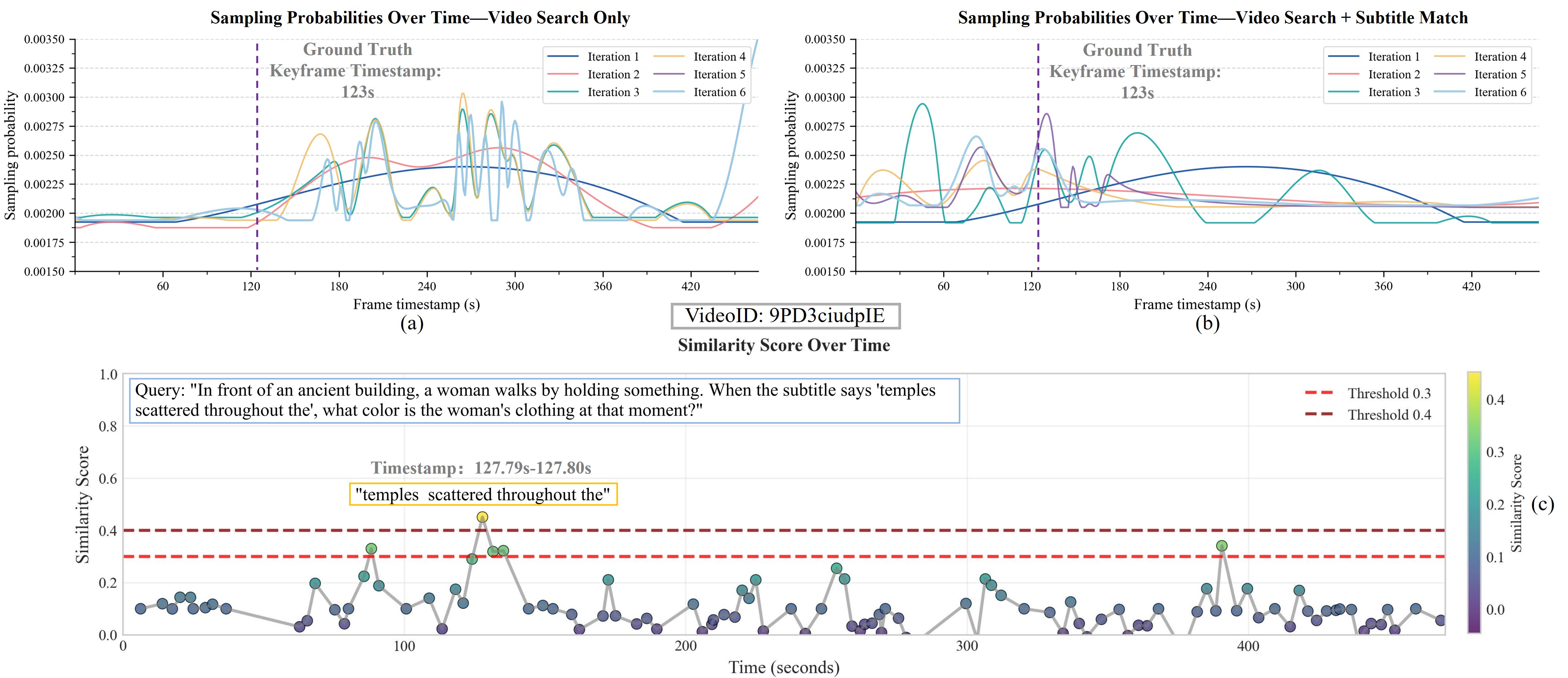}
\caption{\textbf{Case study of branch interaction}. (a) Evolution of the Video Search sampling distribution across iterations. (b) Distribution shift after subtitle-score fusion. (c) The highest-similarity subtitle appears at 127.79s to 127.80s and acts as a temporal cue. Together, these views show how subtitle semantics steer Video Search toward query-relevant regions with stronger temporal consistency and localization stability.}
\label{cs}
\end{figure*}



\begin{algorithm}[t]
\small
\caption{\methods $(V, Q, K, \tau_{\text{obj}})$} 
\begin{algorithmic}[1]
\Require Video $V = \{v_t\}_{t=1}^{N_v}$, Query $Q$, Subtitles $\mathcal{C}$ 
\Ensure Keyframes $\mathcal{K} = \{v_{k_j}\}_{j=1}^K$ 
\State $\mathcal{T} \gets \textsc{VLM}(V, Q, K)$ 
\State $D \gets \mathcal{U}(1,N_v)$, $\mathcal{S}_f \gets \mathbf{0}_{N_v}$, $B_{\text{max}} \gets B_0$ 

\While{$B_{\text{max}} > 0$ \textbf{and} $\neg \textsc{AllFound}(\mathcal{T})$} 
    \State $I_s \gets \textsc{Sampling}(D, m^2)$ 
    \State $S_o \gets \textsc{ObjectScoring}(I_s, \tau_{\text{obj}})$ 
    \State $S_t \gets \textsc{TextSearch}(Q, \mathcal{C})$ 
    \State $S_f \gets \textsc{ScoreFusion}(S_o, S_t, Q)$ 
    \State $\textsc{UpdateDistribution}(D, S_f)$ 
    \State $B_{\text{max}} \gets B_{\text{max}} - |I_s|$ 
\EndWhile

\State \Return $\textsc{TopK}(\mathcal{S}_f, K)$ \hfill {// Final keyframe selection}
\end{algorithmic}
\end{algorithm}
Given an input tuple comprising subtitle text segments $\mathcal{C}=\{c_1, c_2, \ldots, c_n\}$ with corresponding timestamps $T=$ $\{t_1, t_2, \ldots, t_n\}$, and a textual query $Q$, our semantic search pipeline operates through the following computational steps: 
1) \textit{Dual-Text Encoding}: A pre-trained transformer encoder $E_{\text{text}}$ (specifically, \texttt{all-mpnet-base-v2}\footnote{\url{https://huggingface.co/sentence-transformers/all-mpnet-base-v2}}) is used to encode both the textual query and subtitle segments into a shared $d$-dimensional latent space:
\begin{equation}
    \mathbf{q} = E_{\text{text}}(Q) \in \mathbb{R}^d,
\end{equation}
\begin{equation}
    \mathbf{c}_i = E_{\text{text}}(c_i) \in \mathbb{R}^d, \quad \forall i \in [1, n], 
\end{equation}
where $Q$ is the input query, $c_i$ denotes the $i$-th subtitle segment, and $d$ is the embedding dimension determined by the text encoder architecture; and 
2) \textit{Similarity Computation}: We calculate the cosine similarity matrix $\mathbf{M} \in \mathbb{R}^n$ between the query embedding and all subtitle embeddings, each element $\mathbf{M}_i$ is formulated as: 

\begin{equation}
    \mathbf{M}_i=\frac{\mathbf{q} \cdot \mathbf{c}_i}{\|\mathbf{q}\|\left\|\mathbf{c}_i\right\|}, \quad \forall i \in[1, n]. 
\end{equation}

\noindent\textbf{Soft Threshold Enhancement:} We apply soft threshold enhancement to the similarity scores $\mathbf{M}_i$, using an amplification factor $\gamma$ and a threshold $\theta$: 
\begin{equation}
    B_i = \min\left( \mathbf{M}_i + \gamma \cdot \max(\mathbf{M}_i - \theta, 0),\ 1.0 \right),
\end{equation}
where $B_i$ denotes the enhanced similarity score after soft thresholding, $\theta = 0.5$ is the similarity threshold, $\gamma = 2$ is the amplification factor, and $\min(\cdot, 1.0)$ ensures that the score does not exceed 1.0.

\noindent\textbf{Dynamic Gaussian Propagation:} For each segment $i$ with $B_i > \tau_{\text{text}}$ ($\tau_{\text{text}} = 0.2$), we propagate its score using a Gaussian kernel centered at the segment midpoint $c'_i = (b_i + e_i)/2$: 
\begin{equation}
    w_i(t) = B_i \cdot \exp\left(-\frac{(t - c'_i)^2}{2\sigma_i^2}\right),
\end{equation}
where $b_i$ and $e_i$ are the begin and end timestamps of subtitle segment $i$, respectively. The standard deviation $\sigma_i$ controls the spread of the kernel and is defined as $\sigma_i = (e_i - b_i + 2W)/{4}$, where $W = 2$ is the temporal extension radius, and $(e_i - b_i)$ is the original duration of the segment. The variable $t$ denotes a continuous time point within the video timeline.

\noindent\textbf{Max-Confidence Aggregation:} We compute the text confidence score $S_{\text{text}}(t)$ for the frame at time $t$ as the maximum propagated score among all relevant segments:
\begin{equation}
    S_{\text{text}}(t) = \max_{i: B_i > \tau_{\text{text}},\, t \in [b_i - W,\ e_i + W]} w_i(t), 
\end{equation}
With the convention that $\max(\emptyset) = 0$, i.e., when no matching segments are found, the score defaults to 0. The aggregation operation under this agreement effectively integrates textual semantic information from adjacent segments across the temporal dimension, ultimately focusing on and extracting the strongest textual signals to achieve precise aggregation of key textual information at the temporal level.

\subsubsection{Score Fusion Module}

We fuse the object confidence score $S_{\text{obj}}(t) \in [0,1]$ and the text confidence score $S_{\text{text}}(t) \in [0,1]$ using an adaptive weighted scheme. The fused score is computed as:
\begin{equation}
    S_{\text{fused}}(t) = \lambda \cdot \mathcal{N}\left(S_{\text{text}}(t)\right) + (1-\lambda) \cdot \mathcal{N}\left(S_{\text{obj}}(t)\right),\label{eq:score_fusion}
\end{equation}
where $\lambda$ is the modality weighting coefficient (specifically the Text Weight), and $\mathcal{N}(x) = (x - \mu_x)/(\sigma_x + \epsilon)$ denotes the Z-score normalization, with $\mu_x$ and $\sigma_x$ being the temporal mean and standard deviation of score stream $x$. Introduce an infinitesimal constant $\epsilon = 1 \times 10^{-6}$ to ensure stability.

\subsubsection{Score Distribution Update}

After each sampling iteration, we update the global score distribution $\{ \mathcal{S}_f \}$ over all frames $f \in \{1, \cdots, N_{\text{video}} \}$. When a frame $f$ is selected for detection, its score is assigned:
\begin{equation}
\mathcal{S}_f \gets \operatorname{ConfidenceScore}(f), \quad \mathcal{V} \gets \mathcal{V} \cup \{f\}. 
\end{equation}

We then reconstruct scores for unvisited frames using spline interpolation over visited frames:
\begin{equation}
\hat{\mathcal{S}}(f) = \operatorname{Spline}\left( \left\{ (f_i, \mathcal{S}_{f_i}) \mid f_i \in \mathcal{V} \right\} \right)(f). 
\end{equation}

This method can generate smooth probability curves that align with the continuity of video content. To avoid degenerate probabilities, we apply a lower-bound correction:
\begin{equation}
\tilde{\mathcal{S}}(f) = \max\left(N_{\text{video}}^{-1}, \hat{\mathcal{S}}(f)\right). 
\end{equation}

Finally, the frame selection distribution is obtained via a normalized sigmoid transformation:
\vspace{-0.55em}
\begin{equation}
P_f = \frac{\sigma\left( \tilde{\mathcal{S}}(f) \right)}{\sum_{f'=1}^{N_{\text{video}}} \sigma\left( \tilde{\mathcal{S}}(f') \right)}. 
\end{equation}
\vspace{-0.7em}

This process is repeated until the computational budget $B_{\text{max}}$ is reached or target and cue objects are detected. The algorithm returns the top-$k$ frames ranked by $\mathcal{S}_f$. 

In summary, the \methods framework provides a novel and effective solution for multimodal video keyframe retrieval by fully integrating the complementary characteristics of visual and textual modalities. To further validate the superiority of the proposed sampling strategy, this paper additionally conducts a comparative experiment, benchmarking the proposed adaptive iterative sampling strategy against the classic \textbf{Epsilon-Greedy}~\citep{e} and \textbf{UCB}~\citep{ucb} strategies. Experimental results demonstrate that compared to these two sampling strategies, the proposed adaptive iterative sampling strategy exhibits significant advantages: not only does it feature lower computational complexity and require fewer iterations per keyframe search, but it also achieves superior overall search efficiency. This outcome further validates the reliability and robustness of the proposed adaptive iterative sampling strategy. Detailed experimental data are in the Appendix.

\section{Experiments}
\begin{table*}[t]
    \setlength\tabcolsep{4pt} 
    \setlength\extrarowheight{1pt} 
    \arrayrulecolor[gray]{0.7} 
    \caption{
\textbf{Downstream QA task evaluation results on \lvb and \videomme}. \textbf{Frame} indicates the number of frames used in performing VideoQA tasks. \textbf{Searching Modality} indicates the primary information modalities utilized by keyframe search methods.
}
\vspace{-0.7em}
\begin{adjustbox}{width=\linewidth}
\begin{tabular}{lcccclcccc}
\hline
\multicolumn{5}{c}{\textbf{\lvb}}                                           & \multicolumn{5}{c}{\textbf{\videomme}}                                                 \\ \hline
\multirow{2}{*}{\textbf{Model}} &
  \multirow{2}{*}{\textbf{Searching Modality}} &
  \multirow{2}{*}{\textbf{Frame}} &
  \multicolumn{2}{c}{\textbf{Video Length}} &
  \multirow{2}{*}{\textbf{Model}} &
  \multirow{2}{*}{\textbf{Searching Modality}} &
  \multirow{2}{*}{\textbf{Frame}} &
  \multicolumn{2}{c}{\textbf{Video Length}} \\ \cline{4-5} \cline{9-10} 
                                 &            &    & Long & Medium &                                  &            &    & Long & Medium \\ \hline
GPT-4O~\citep{hurst2024gpt}                           & N/A        & 8  & 47.0      & 50.5        & GPT-4O                           & N/A        & 8  & 54.5      & 61.2        \\
GPT-4O + Tstar~\citep{ye2025rethinking}                  & Unimodal   & 8  & 47.3      & 52.9        & GPT-4O + Tstar                   & Unimodal   & 8  & 55.0      & 59.7        \\
GPT-4O + VSLS~\citep{vsls}                    & Unimodal   & 8  & 49.1      & 52.2        & GPT-4O + VSLS                    & Unimodal   & 8  & 55.6      & 61.1        \\
\rowcolor{gray!10}
GPT-4O + \methods (ours)              & Multimodal & 8  & \textbf{49.2}      & \textbf{56.1}        & GPT-4O + \methods (ours)              & Multimodal & 8  & \textbf{55.8}      & 59.5        \\
LLaVA-Video-7B-Qwen2~\citep{36}             & N/A        & 8  & 39.0      & 42.7        & LLaVA-Video-7B-Qwen2             & N/A        & 8  & 38.0      & 39.7        \\
LLaVA-Video-7B-Qwen2+Tstar       & Unimodal   & 8  & 40.8      & 44.1        & LLaVA-Video-7B-Qwen2+Tstar       & Unimodal   & 8  & 37.5      & 40.4        \\
LLaVA-Video-7B-Qwen2+VSLS        & Unimodal   & 8  & 38.7      & 42.4        & LLaVA-Video-7B-Qwen2+VSLS        & Unimodal   & 8  & 38.5      & 38.5        \\
\rowcolor{gray!10}
LLaVA-Video-7B-Qwen2+\methods(ours)   & Multimodal & 8  & \underline{40.6}      & \textbf{44.1}        & LLaVA-Video-7B-Qwen2+\methods(ours)   & Multimodal & 8  & \textbf{38.8}      & \textbf{42.2}        \\
QWEN2.5-VL-7B-INSTRUCT~\citep{qwen}           & N/A        & 8  & 38.7      & 43.5        & QWEN2.5-VL-7B-INSTRUCT           & N/A        & 8  & 38.0      & 47.3        \\
QWEN2.5-VL-7B-INSTRUCT+Tstar     & Unimodal   & 8  & 39.2      & 43.9        & QWEN2.5-VL-7B-INSTRUCT+Tstar     & Unimodal   & 8  & 40.3      & 50.0        \\
QWEN2.5-VL-7B-INSTRUCT+VSLS      & Unimodal   & 8  & 42.4      & 43.9        & QWEN2.5-VL-7B-INSTRUCT+VSLS      & Unimodal   & 8  & 43.2      & 49.6        \\
\rowcolor{gray!10}
QWEN2.5-VL-7B-INSTRUCT+\methods(ours) & Multimodal & 8  & \underline{39.9}& \textbf{45.1}        & QWEN2.5-VL-7B-INSTRUCT+\methods(ours) & Multimodal & 8  & \underline{40.8}      & 48.5        \\ \hline
GPT-4O                           & N/A        & 32 & 48.2      & 51.9        & GPT-4O                           & N/A        & 32 & 55.2      & 61.0        \\
GPT-4O + Tstar                   & Unimodal   & 32 & 46.6      & 52.5        & GPT-4O + Tstar                   & Unimodal   & 32 & 55.2      & 61.6        \\
GPT-4O + VSLS                    & Unimodal   & 32 & 49.1      & 50.7        & GPT-4O + VSLS                    & Unimodal   & 32 & 57.5      & 61.9        \\
\rowcolor{gray!10}
GPT-4O + \methods (ours)              & Multimodal & 32 & \textbf{51.2}      & \textbf{53.9}        & GPT-4O + \methods (ours)              & Multimodal & 32 & \underline{55.3}      & \underline{61.7}        \\
LLaVA-Video-7B-Qwen2             & N/A        & 32 & 39.4      & 45.3        & LLaVA-Video-7B-Qwen2             & N/A        & 32 & 35.9      & 36.5        \\
LLaVA-Video-7B-Qwen2+Tstar       & Unimodal   & 32 & 39.2      & 43.9        & LLaVA-Video-7B-Qwen2+Tstar       & Unimodal   & 32 & 35.8      & 39.6        \\
LLaVA-Video-7B-Qwen2+VSLS        & Unimodal   & 32 & 42.4      & 43.9        & LLaVA-Video-7B-Qwen2+VSLS        & Unimodal   & 32 & 36.9      & 39.0        \\
\rowcolor{gray!10}
LLaVA-Video-7B-Qwen2+\methods(ours)   & Multimodal & 32 & \underline{39.9} & \textbf{45.3}        & LLaVA-Video-7B-Qwen2+\methods(ours)   & Multimodal & 32 & \textbf{38.2}      & \textbf{42.2}        \\
QWEN2.5-VL-7B-INSTRUCT           & N/A        & 32 & 30.3      & 35.0        & QWEN2.5-VL-7B-INSTRUCT           & N/A        & 32 & 37.5      & 39.9        \\
QWEN2.5-VL-7B-INSTRUCT+Tstar     & Unimodal   & 32 & 38.7      & 42.9        & QWEN2.5-VL-7B-INSTRUCT+Tstar     & Unimodal   & 32 & 34.9      & 45.6        \\
QWEN2.5-VL-7B-INSTRUCT+VSLS      & Unimodal   & 32 & 38.1      & 43.1        & QWEN2.5-VL-7B-INSTRUCT+VSLS      & Unimodal   & 32 & 37.9      & 50.0        \\
\rowcolor{gray!10}
QWEN2.5-VL-7B-INSTRUCT+\methods(ours) & Multimodal & 32 & 37.1      & \textbf{44.4}        & QWEN2.5-VL-7B-INSTRUCT+\methods(ours) & Multimodal & 32 & \textbf{39.3}      & \underline{45.8}        \\ \hline
\end{tabular}
\end{adjustbox}
    \label{VideoQA}
    \vspace{-0.7em}
\end{table*}
\subsection{Benchmark Datasets}
We evaluate \methods on two representative benchmarks: \lvb{} (LVB)~\citep{lvb} and \videomme{} (VMME)~\citep{videomme}. LVB provides ground-truth keyframe annotations for direct evaluation of retrieval quality, while both benchmarks support downstream VideoQA assessment. Detailed dataset statistics are provided in the Appendix.

\subsection{Evaluation Metrics for Search Utility}
Search utility is evaluated from two complementary perspectives.
First, \textbf{Acc} measures coarse temporal localization utility: a retrieval is considered valid if at least one frame in the top-k results lies within 200 frames of the ground-truth keyframe index. Acc is computed as the ratio of valid retrievals to all evaluated videos.

Second, we measure fine-grained frame-level utility using SSIM-based set matching. Specifically, \textbf{SSIM-Precision} measures the quality (purity) of predicted keyframes, while \textbf{SSIM-Recall} measures the coverage of ground-truth keyframes. These metrics complement Acc by capturing structural fidelity beyond hit-or-miss temporal matching.

Let the predicted and ground-truth keyframe sets be
$\mathcal{K}_{\text{pred}} = \{k_i^{\text{pred}}\}_{i=1}^{M}$ and $\mathcal{K}_{\text{gt}} = \{k_j^{\text{gt}}\}_{j=1}^{N}$, where $I(\cdot)$ maps a frame index to its image. We define pairwise similarity as
$\rho_{ij} = \operatorname{SSIM}(I(k_i^{\text{pred}}), I(k_j^{\text{gt}}))$~\cite{ssim}. Then,
\vskip -0.5em
\begin{subequations}
\begin{empheq}[left=\empheqlbrace]{align}
\operatorname{Prec}(\mathcal{K}_{\text{pred}}, \mathcal{K}_{\text{gt}}) &= \frac{1}{M} \sum_{i=1}^{M} \max_{1 \le j \le N} \rho_{ij},\label{eq:ssim_precision} \\
\operatorname{Rec}(\mathcal{K}_{\text{pred}}, \mathcal{K}_{\text{gt}}) &= \frac{1}{N} \sum_{j=1}^{N} \max_{1 \le i \le M} \rho_{ij}. \label{eq:ssim_recall}
\end{empheq}
\end{subequations}

\subsection{Evaluation Metrics for Search Efficiency}
For search efficiency, we report three metrics: \textbf{TFLOPs}, \textbf{Iteration}, and \textbf{Latency(sec)}. These metrics jointly reflect computational cost, average search steps, and end-to-end runtime in practical long-video retrieval scenarios.

We compare against Tstar and VSLS, two representative keyframe-search baselines. As shown in Table \ref{tab:main_bench_efficiency}, \methods achieves the best retrieval performance on the full \lvb benchmark and on both image-only and text-only subsets. The gain is most evident in text-centric settings, indicating that subtitle cues provide complementary guidance to visual search. At the same time, \methods maintains competitive efficiency in TFLOPs, iteration count, and latency, yielding a favorable effectiveness-efficiency trade-off.

\begin{table}[t]
  \centering
  \footnotesize            
  \setlength{\tabcolsep}{3pt}
  \setlength\extrarowheight{1pt}

  \caption{\small{Search utility results on \lvb. VSI achieves the best SSIM-Recall and Acc.}}
  \vspace{-0.5em}
  \label{tab:framesearch_evaluation}

  \begin{adjustbox}{width=\linewidth}
  \begin{tabular}{lcccc}
  \toprule
  \multicolumn{1}{l|}{\multirow{2}{*}{\textbf{Method}}} &
  \multicolumn{1}{c|}{\multirow{2}{*}{\textbf{Frame}}} &
  \multicolumn{3}{c}{\textbf{\lvb}} \\
  \multicolumn{1}{l|}{} & \multicolumn{1}{c|}{} &
  Precision $\uparrow$ & Recall $\uparrow$ & Acc $\uparrow$ \\ \hline
  \multicolumn{5}{c}{Static Frame Sampling Method} \\ \hline
  
  \textsc{Uniform}                                    & 8  & 60.7 & 80.4 & - \\ \hline
  
  \multicolumn{5}{c}{Dense Retrieval Method} \\ \hline
  \textsc{VideoAgent}~\cite{13}           & 10.1 & 58.8 & 73.2 & - \\
  \textsc{Retrieval-based}~\cite{ye2025rethinking} & 8 & 63.1 & 65.5 & - \\ \hline
  \multicolumn{5}{c}{Temporal Searching Method} \\ \hline
  Tstar~\cite{ye2025rethinking}                                             & 8  & 75.3 & 88.2 & 37.1  \\
  
  VSLS~\cite{vsls}                & 8  & 75.6 & 88.6 & 37.0 \\
  AKS~\cite{aks}               & 8  & 78.0 & 83.0 & 36.6 \\
  \rowcolor{gray!10}
  VSI(ours)           & 8  & \underline{76.8} & \textbf{89.7} & \textbf{37.9} \\
  
  \bottomrule
  \end{tabular}
  \end{adjustbox}
\end{table}

\subsection{Downstream VideoQA Performance}
\begin{table*}[t]
    \centering
    \setlength\tabcolsep{8pt}
    \setlength\extrarowheight{1pt}
\caption{Evaluation of computational performance metrics across the \lvb benchmark. \textbf{Training Required} shows whether the method needs to be trained, \textbf{Video Search} indicates the model used by the method for video modality processing, and \textbf{Text Encoding} indicates the model used by the method for text modality processing.}
\vspace{-0.7em}
\begin{adjustbox}{width=\linewidth}
    
\begin{tabular}{lcccccccc}
\toprule
\multicolumn{1}{c}{} &               & \multicolumn{5}{c}{\textbf{Searching Efficiency}}                   &          &       \\ \cline{3-7}
\multicolumn{1}{c}{\multirow{-2}{*}{\textbf{Method}}} &
  \multirow{-2}{*}{\textbf{Training Required}} &
  Video Search &
  Text Encoding &
  TFLOPs &
  Iteration &
  Latency(sec) &
  \multirow{-2}{*}{\textbf{Frame}} &
  \multirow{-2}{*}{\textbf{Acc} $\uparrow$} \\ \hline
\multicolumn{9}{c}{ALL}                                                                      \\ \hline
T*-DETECTOR~\citep{ye2025rethinking}          & Training Free & YOLO-World-110M & N/A               & 31.7 & 24.76 & 28.96 & 64       & 67.58 \\
VSLS-DETECTOR~\citep{vsls}        & Training Free & YOLO-World-110M & N/A               & 33.3 & 24.49 & 33.26 & 64       & 70.23 \\
\rowcolor{gray!10}
\methods-DETECTOR(ours)         & Training Free & YOLO-World-110M & ALL-MPNET-BASE-V2 & 36.8 & 26.34 & 31.71 & 64       & \textbf{73.89} \\ \hline
\multicolumn{9}{c}{IMAGE-ONLY}                                                               \\ \hline
T*-DETECTOR          & Training Free & YOLO-World-110M & N/A               & 31.7 & 23.90 & 25.4  & 64       & 70.34 \\
VSLS-DETECTOR        & Training Free & YOLO-World-110M & N/A               & 33.3 & 24.04 & 28.13 & 64 & 68.62 \\
\rowcolor{gray!10}
\methods-DETECTOR(ours)         & Training Free & YOLO-World-110M & ALL-MPNET-BASE-V2 & 36.8 & 25.35 & 30.92 & 64       & \textbf{71.56} \\ \hline
\multicolumn{9}{c}{TEXT-ONLY}                                                                \\ \hline
T*-DETECTOR          & Training Free & YOLO-World-110M & N/A               & 31.7 & 26.80 & 27.99 & 64       & 66.36 \\
VSLS-DETECTOR        & Training Free & YOLO-World-110M & N/A               & 33.3 & 26.39 & 28.77 & 64       & 66.36 \\
\rowcolor{gray!10}
\methods-DETECTOR(ours)         & Training Free & YOLO-World-110M & ALL-MPNET-BASE-V2 & 36.8 & 28.37 & 34.44 & 64       & \textbf{77.17} \\ 
\bottomrule
\end{tabular}
\end{adjustbox}
    \label{tab:main_bench_efficiency}
    \vspace{-1em}
\end{table*}
We further evaluate downstream VideoQA on both \lvb and \videomme, focusing on \textbf{Medium} (LVB: 3-10 min, VMME: 4-15 min) and \textbf{Long} (LVB: 10-60 min, VMME: 30-60 min) videos.

Using GPT-4o, LLaVA-Video-7B-Qwen2, and Qwen2.5-VL-7B-instruct as foundation models, we test two frame budgets (8 and 32). Table \ref{VideoQA} shows that \methods consistently improves over uniform-sampling baselines on \lvb and remains competitive on \videomme, achieving state-of-the-art performance on medium-length LVB tasks and demonstrating strong generalization to medium-to-long video QA.

\subsection{Visual Subtitle Integration Performance}
To validate the contribution of subtitle-aware reasoning, we further evaluate \methods on the text-related perception subset of \lvb, which includes Text-referred Object Attribute (T2A), Text-referred Event (T2E), and Text-referred Object (T2O). These tasks require tight alignment between visual and textual signals, making them a suitable testbed for multimodal keyframe retrieval. As shown in Table \ref{ablation_text_ptonly}, \methods consistently improves over unimodal baselines in keyframe search and long-video QA, showing that subtitle information provides effective complementary guidance for long-video understanding. We also compare against the subtitle-augmented uniform-sampling baseline (GPT-4o+Sub). \methods remains superior on long-video QA, suggesting that the improvement comes from explicit visual-textual integration during frame selection rather than from simply adding subtitle inputs in a prompt-only manner.

Consistent with the metric definitions above, Table~\ref{tab:framesearch_evaluation} evaluates search utility from visual similarity (Precision and Recall) and coarse temporal localization (Acc). Under the 8-frame setting, \methods achieves the best Recall (89.7) and the best Acc (37.9\%), while maintaining competitive Precision (76.8). Compared with strong temporal-search baselines, \methods improves Recall by 1.1 over VSLS (88.6) and improves Acc by 0.8\% over Tstar (37.1). Although AKS reports slightly higher Precision (78.0), its lower Recall and Acc indicate weaker coverage of query-relevant content. These results suggest that subtitle-guided visual search improves temporal completeness while preserving strong visual-semantic alignment across diverse query intents.
\begin{table}[t]
    \centering
    \footnotesize
    \caption{Keyframe Search Accuracy and Downstream QA Accuracy on \lvb-Text-Related Perception Subsets.}
    \vspace{-0.7em}
    \begin{adjustbox}{width=\linewidth}
    \begin{tabular}{lcccc}  
        \toprule
        \multirow{2}{*}{\textbf{Method}}  & \multirow{2}{*}{\textbf{Frame}} &  \multirow{2}{*}{\textbf{Acc} $\uparrow$} &
        \multicolumn{2}{c}{\textbf{\lvb}} \\  
        \cmidrule(lr){4-5}  
         & &  & Medium & Long \\  
        \midrule
        \textsc{GPT4o}~\citep{lvb}  &8 & N/A & 48.28 & 53.76 \\
        \textsc{GPT4o+Sub}~\citep{lvb}  &8 & N/A & 46.55 & 58.06 \\
        \textsc{GPT4o+Tstar}~\citep{ye2025rethinking}  &8 & 29.48 & 48.28 & 58.06 \\
        \textsc{GPT4o+VSLS}~\citep{vsls}  &8 & 27.17 & 48.28 & 55.91 \\
        \rowcolor{gray!10}
        \textsc{\textbf{GPT4o+\methods}}~(ours)  &8 & \textbf{45.00} & \textbf{62.07} & \textbf{69.57} \\
        \bottomrule
    \end{tabular}
    \end{adjustbox}
    \label{ablation_text_ptonly}
    \vspace{-2.2em}
\end{table}
\vspace{-0.7em}
\section{Analysis}
\subsection{Video Search and Subtitle Match}
To quantify how each branch contributes to retrieval quality, we compare three settings in Table \ref{tab:branch_ablation}: Subtitle Match Only, Video Search Only, and Fusion (32-frame setting). The standalone results show complementary strengths: Subtitle Match favors Recall (94.1), while Video Search favors Precision (75.0). Their fusion achieves the best overall trade-off, yielding the highest F1 Score (83.7) together with the highest Precision (75.2) and Recall (94.3). These results indicate that neither branch is redundant, and both are necessary for robust global performance.

From the perspective of branch interaction, the two branches are complementary but asymmetric. During iterative keyframe search, Video Search serves as the primary retrieval engine. Subtitle Match acts as an auxiliary branch by injecting text semantics into the Video Search sampling distribution through score fusion, which helps concentrate probability mass on segments containing core query semantics. In final keyframe selection, the unified score distribution combines signals from both branches, and each branch's effective contribution is adaptively determined by whether the query is more image-dominant or text-dominant.

Figure \ref{cs} provides a concrete example. The subtitle with the highest semantic similarity appears at 127.79s to 127.80s. After score fusion, this temporal cue is propagated to the Video Search distribution, so subsequent iterations concentrate more precisely on visually relevant regions. This behavior directly demonstrates how text-modal signals guide visual search and improve cross-modal consistency throughout iterative retrieval cycles on semantically dense videos.
\begin{table}[t]
\centering
\caption{Independent and joint contributions of the two branches on \lvb (32-frame setting). Subtitle Match Only favors Recall, Video Search Only favors Precision, and their Fusion achieves the best overall trade-off, yielding the highest PRF Score.}
\vspace{-0.5em}
\label{tab:branch_ablation}
\resizebox{\linewidth}{!}{
\begin{tabular}{l|ccc}
\toprule
\textbf{Setting} & \textbf{Precision} $\uparrow$ & \textbf{Recall} $\uparrow$ & \textbf{F1 Score} $\uparrow$ \\
\hline
VSI (Subtitle Match Only) & 74.3 & 94.1 & 83.0 \\
\rowcolor{gray!10}
VSI (Video Search Only) & 75.0 & 93.5 & 83.2 \\
\rowcolor{gray!20}
VSI (Fusion) & \textbf{75.2} & \textbf{94.3} & \textbf{83.7} \\
\bottomrule
\end{tabular}
}
\vspace{-1.5em}
\end{table}

\subsection{Subtitle Dependency and Robustness}
The dual-branch architecture is not a simple summation but a collaborative mechanism in which the Subtitle branch provides coarse temporal priors and the Visual branch performs fine-grained frame localization. As shown in Table~\ref{tab:subtitle_robustness}, VSI degrades gracefully under subtitle corruption. Compared with the standard setting (Precision/Recall/F1 = 75.2/94.3/83.7), the No Subtitle setting changes to 75.0/93.5/83.2, and the Noisy Subtitle setting changes to 74.6/92.7/82.7. These results indicate that noisy or missing subtitles introduce limited distraction but do not cause system collapse. Importantly, both degraded settings remain above visual-only baselines (VSLS: 82.5 F1; Tstar: 81.3 F1), confirming that subtitle guidance is a safe enhancement with robust fallback behavior.

\begin{table}[h]
\centering
\vspace{-0.5em}
\caption{Robustness of \methods{} under subtitle degradation on \lvb{} (32-frame setting). \methods remains stable when subtitles are missing or noisy and consistently outperforms visual-only baselines, confirming robust fallback behavior.}
\vspace{-0.5em}
\label{tab:subtitle_robustness}
\resizebox{\linewidth}{!}{
\begin{tabular}{l|c|ccc}
\toprule
\textbf{Method} & \textbf{Subtitle Availability} & \textbf{Precision} $\uparrow$ & \textbf{Recall} $\uparrow$ & \textbf{F1 Score} $\uparrow$ \\
\hline
VSI (Standard) & 100\%  Subtitles & \textbf{75.2} & \textbf{94.3} & \textbf{83.7} \\
VSI (No Subtitle) & 0\% Subtitles & 75.0 & 93.5 & 83.2 \\
VSI (Noisy Subtitle) & Noisy Subtitles & 74.6 & 92.7 & 82.7 \\
\hline
VSLS (Baseline) & N/A (Visual Only) & 74.5 & 92.5 & 82.5 \\
Tstar (Baseline) & N/A (Visual Only) & 74.0 & 90.3 & 81.3 \\
\bottomrule
\end{tabular}
}
\vspace{-1.5em}
\end{table}

\subsection{Complementarity and Overlap Analysis}
To address concerns about redundancy and conflict between the Video Search and Subtitle Match branches, we quantitatively analyze the overlap rate of the top-K keyframes selected by the two standalone branches. The average overlap rate is only 34.60\%, indicating a high degree of complementarity between the two modalities. This means that the dual-branch design enables VSI to cover 65.40\% of key information points that would otherwise be missed by single-modality retrieval.

Notably, our fusion framework does not maintain separate candidate lists for the two branches. Instead, it directly fuses branch scores to generate a unified sampling probability distribution. Frames with high scores in both modalities naturally receive higher sampling probabilities because they satisfy both visual and semantic relevance. When the two branches assign high scores to different frames, priority is determined by the adaptive modality weighting coefficients in Eq.~\ref{eq:score_fusion}, which balance visual and textual evidence according to task type. For text-dominated tasks, VSI assigns a higher effective weight to Subtitle Match, while for vision-dominated tasks, it prioritizes Video Search.

\subsection{Time Complexity}
The proposed multimodal framework extends the unimodal approach by incorporating textual semantic analysis alongside visual processing. The enhanced method maintains the original two-stage architecture while introducing a parallel text processing pipeline. The multimodal search retains the original $\mathcal{O}(|\mathcal{S}|k \log n)$ YOLO-World detection complexity, with two key additions: 1) Text Processing Overhead: subtitle analysis introduces $\mathcal{O}(m)$ complexity for $m$ text segments, which is negligible compared with visual processing; 2) Fusion Cost: cross-modal scoring adds a constant $\mathcal{O}(1)$ overhead per frame. YOLO-World detection remains the computational bottleneck. Our implementation shows only 10.5\% increased inference time versus the visual-only baseline, demonstrating efficient multimodal integration.

\section{Related work}
\noindent \textbf{Challenges in Long-Form Video Understanding.} Long-form video analysis~\citep{xun2025rtv,liu2025javisgpt} poses greater and more persistent challenges than short-video or image tasks~\citep{survey}, primarily due to inherently complex temporal dynamics and substantial content redundancy~\citep{qian2024streaming, zeng2024timesuite, yu2019activityqa,dang2024exploring}. The large number of frames introduces significant computational burdens, substantially increasing memory and latency, making exhaustive frame-wise processing infeasible. Moreover, key semantic events are often temporally sparse and discontinuous, requiring specialized architectures capable of capturing subtle, long-range dependencies~\citep{ranasinghe2025understanding, shi2024unlocking, chen2024rextime,weng2024longvlm}. The visual complexity of such videos also introduces noisy, ambiguous, or irrelevant signals, which calls for more effective content distillation mechanisms~\citep{84, cheng2024videollama2advancingspatialtemporal, xu2023retrieval, ye2025rethinking,hjx,fmri,d}.

\noindent \textbf{Video-Language Modeling Strategies.} Recent work using VLMs has primarily focused on three complementary directions: 1) content-aware sampling for more efficient computation~\citep{buch2025flexible,yu2025framevoyager,hu2025mllm,tspo,moss}, 2) hierarchical reasoning for complex query understanding~\citep{wang2024videoagent,min2024morevqa,liao2024align}, and 3) information compression to better mitigate context window limitations~\citep{tao2025dycoke,liu2025hicom,choudhury2024rlt,chen2025longvila}. Representative strategies include segmentation-retrieval pipelines that divide long videos into clips and employ either learned or heuristic mechanisms to reliably select relevant segments~\citep{pan2023retrieving, choudhury2023zero, rohan2025video}, and fine-grained frame-level token compression to significantly reduce computational costs~\citep{li2024llamavid, chen2024llavolta, song2024less}, or adopt memory propagation techniques to better maintain temporal continuity~\citep{qian2024streaming, wu2022memvit, liu2024llava}. Recent advances also explore planning systems that decompose complex tasks into sequential perception steps, thereby enabling dynamic and adaptive frame-level information acquisition~\citep{wang2024videoagent, liao2024videoinsta}.

\section{Conclusion}

This paper proposes \method(\methods), a multimodal keyframe search framework built on a dual-branch cross-modal retrieval mechanism that combines Video Search and Subtitle Match. On \lvb, VSI achieves strong retrieval effectiveness with 73.89\% Acc and improves downstream VideoQA over uniform sampling baselines. Beyond headline performance, our analyses clarify why the method works: branch ablation shows that fusion yields the best overall trade-off, SSIM-based evaluation under the 8-frame setting reaches the best Recall and best Acc with competitive Precision, robustness tests show graceful degradation under missing or noisy subtitles while remaining above visual-only baselines, and the low cross-branch overlap rate confirms strong cross-modal complementarity. Together with only 10.5\% additional runtime over visual-only search, these results demonstrate that VSI is an efficient and plug-and-play solution for long-video understanding without extra training or annotation.

\newpage
{\small
\bibliographystyle{ieee_fullname}
\bibliography{egbib}
}

\end{document}